\documentclass[letterpaper, 10 pt, conference]{ieeeconf}  

\IEEEoverridecommandlockouts                              

\usepackage{cite}
\usepackage{amsmath,amssymb,amsfonts}
\usepackage{algorithmic}
\usepackage{graphicx}
\usepackage{textcomp}
\usepackage{xcolor}
\usepackage{balance}
\usepackage{multirow}
\usepackage{capt-of}
\usepackage[table]{colortbl}
\usepackage{booktabs}
\usepackage{amssymb}
\usepackage{nameref}
\usepackage{xcolor}
\usepackage{etoolbox}
\usepackage{comment}
\usepackage{hyperref}
\let\labelindent\relax
\usepackage{enumitem}
\usepackage{tikz}

\newcommand{\pquotes}[1]{\textcolor[gray]{0.25}{\textit{#1}}}

\newcommand{\highlight}[2]{\colorbox{#1}{#2}}
\definecolor{lightblue}{HTML}{A4E4FF}

\definecolor{customblue}{HTML}{0081B8}

\def\eg{\emph{e.g., }} 
\def\ie{\emph{i.e., }} 

\title{\LARGE \bf
Robot Error Awareness Through Human Reactions:\\ Implementation, Evaluation, and Recommendations}

\author{Maia Stiber$^{1,\dag}$, Russell Taylor$^{1,2}$, and Chien-Ming Huang$^{1}$
\thanks{*This work was supported by NSF award \#2143704. This work has been submitted to the IEEE for possible publication. Copyright may be transferred without notice, after which this version may no longer be accessible.}%
\thanks{$^{1}$Dept. of Computer Science, Johns Hopkins University, Baltimore, Maryland, USA 
        {\tt\small \{mstiber,rht,chienming.huang\}@jhu.edu}}%
\thanks{$^{2}$ Life Fellow, IEEE $^{\dag}$ Corresponding Author}%
\thanks{\textbf{AI Statement.} This paper has been proofread by a language model (AI), and the authors have read through the resulting content to ensure it accurately reflects the original intent.}
\thanks{\textbf{CRediT author statement}. MS: Conceptualization, Methodology, Software, Validation, Formal analysis, Investigation, Data curation, Writing, Visualization. RT: Conceptualization, Methodology, Writing-Review\&editing, Supervision. CMH: Conceptualization, Methodology, Writing-Review\&editing, Supervision, Visualization, Funding acquisition.}
}

\usepackage{etoolbox}
\newcommand{\insertfig}{\vspace{1em}\includegraphics[width=\linewidth]{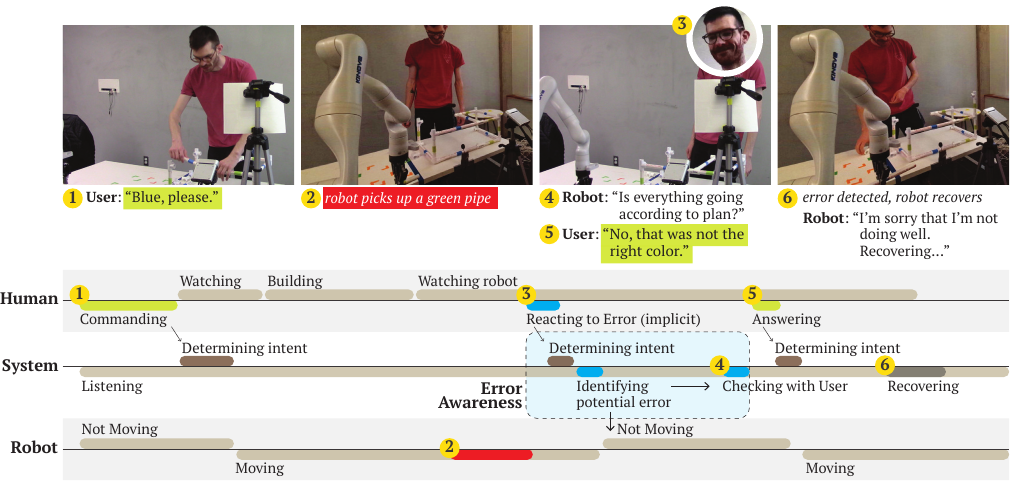}\captionof{figure}{We introduce a proactive robot error detection system that uses human reactions (speech and facial action units) to detect errors in real-time and demonstrate the system's reliability and flexibility across tasks and participants.}\label{fig:teaser}}

\makeatletter
\apptocmd{\@maketitle}{\centering\setcounter{figure}{0}\insertfig}{}{}
\makeatother

\begin{document}

\maketitle
\thispagestyle{empty}
\pagestyle{empty}

\begin{abstract}
Effective error detection is crucial to prevent task disruption and maintain user trust. Traditional methods often rely on task-specific models or user reporting, which can be inflexible or slow. Recent research suggests social signals, naturally exhibited by users in response to robot errors, can enable more flexible, timely error detection. However, most studies rely on post hoc analysis, leaving their real-time effectiveness uncertain and lacking user-centric evaluation. In this work, we developed a proactive error detection system that combines user behavioral signals (facial action units and speech), user feedback, and error context for automatic error detection. In a study ($N = 28$), we compared our proactive system to a status quo reactive approach. Results show our system 1) reliably and flexibly detects error, 2) detects errors faster than the reactive approach, and 3) is perceived more favorably by users than the reactive one. 
We discuss recommendations for enabling robot error awareness in future HRI systems.
\end{abstract}


\section{Introduction}
Robot errors can severely disrupt tasks, degrade user trust, and negatively impact the overall perception of the robot~\cite{helmreich2000error,salem2015would}. 
In human-robot collaborations, where task performance and mutual trust are crucial~\cite{dinh2017factors}, 
unmanaged errors can lead to user frustration and reluctance to continue working with the robot~\cite{hancock2011meta}. Therefore, timely detection and effective management of these errors are essential for maintaining successful collaboration and task execution~\cite{yasuda2013psychological}.

Status quo error detection methods in HRI are often \textit{rigid}, if dependent on task- or domain-specific information for detection models~\cite{chance2016assistive,xu2024sedmamba} and/or \textit{reactive}, if relying on user reports~\cite{edirisinghe2024field}. They typically require manual adjustments for different tasks or error types in the case of task model-based approaches~\cite{honig2021expect} or they delay error detection until a user explicitly reports a problem.

One way to overcome these limitations is by leveraging implicit social signals to detect potential robot errors~\cite{mirnig2017to,stiber2023using}. 
This is promising because people naturally react to unexpected events, such as errors, with social signals. These signals could enable \textit{flexible}, \textit{timely} error detection, potentially identifying issues before users explicitly report them~\cite{ stiber2023using,kontogiorgos2021systematic}. 
Despite this potential, no system to our knowledge has incorporated social signals for real-time error detection. Research in this area has relied on post hoc analysis, leaving questions about whether this method of using implicit signals can effectively support real-time HRI.

Here we developed an error detection system that uses implicit and explicit social signals combined with error context to automatically detect errors in real-time. Our objectives were to enable flexible, automatic, timely, and \emph{proactive} error detection, while maintaining reliability and supporting natural interactions. Our system achieves proactivity by analyzing implicit user reactions in a multi-phase process, while also integrating explicit error reports to ensure robustness. Fig.~\ref{fig:teaser} provides an example interaction with the system. 

Through an evaluation study ($N = 28$), we found that our \emph{proactive} system reliably managed errors across tasks and participants and detected errors faster than a reactive system, allowing for a controlled comparison between using and not using implicit social signals. Participants rated the robot with our \emph{proactive} system more favorably as a teammate and preferred its error-handling approach compared to a \emph{reactive} system in the context of collaborative assembly.

Our work makes the following contributions: 
\begin{itemize} [leftmargin=*]
    \item  \textbf{Design and implementation}
    of a proactive error detection system that uses implicit multimodal human reactions for effective detection across tasks and people. 
    \item \textbf{Evaluation and demonstration} that our system outperforms a status quo, reactive error detection system in both capabilities and user perception.
    
    \item \textbf{Recommendations}  for enabling error-awareness in future interactive robot systems.
\end{itemize}

\section{Related Work}

Robot errors are inevitable and regularly occur in deployed robots~\cite{nourbakhsh2003mobot, honig2018understanding}. 
In HRI, the perception of a robot's action as an error is influenced by task context and whether the behavior deviates from user expectations~\cite{rossi2017human}. Errors can impair task performance and erode user trust, the degree being dependent on the frequency, severity, and context of the errors~\cite{waveren2019take,gideoni2024personal}.
Ignoring errors is not sustainable, as their negative effects accumulate over time, requiring more effort to repair~\cite{helmreich2000error, salem2015would}. Consequently, deployed robots must manage errors by understanding their impact, implementing mitigation strategies, and preventing future occurrences~\cite{honig2018understanding}.

\subsection{Status Quo Techniques for Error Detection in HRI}
Prior detection techniques commonly involve explicit handling (\eg human manual reporting~\cite{edirisinghe2024field}) and/or task- or domain-specific information (\eg hierarchical action structure~\cite{li2022runtime} and characteristics of anticipated errors~\cite{chance2016assistive})~\cite{honig2018understanding}. 
They generally have error handling policies specifically designed for each task and assume a set of predefined anticipated errors~\cite{honig2021expect}. They overlook factors such as the teammate's values and task context, which can affect user perception of whether actions are errors. This limitation makes it difficult to generalize across people, tasks, and errors~\cite{correia2018exploring,rossi2017human}. However, error detection needs to be adaptable, as robot errors often occur unanticipated ways.

\begin{figure*}
\includegraphics[width = \textwidth]{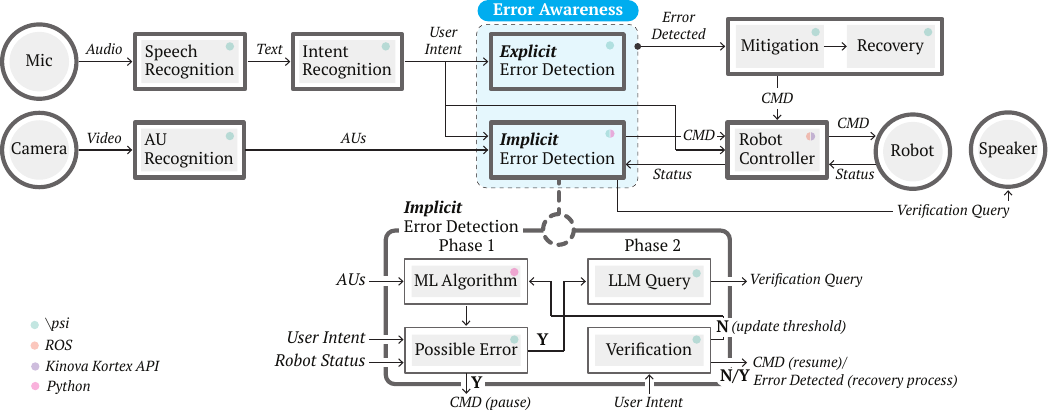}
\caption{System diagram with a zoomed-in view of the implicit error detection component to highlight the two-phased detection process based on the implicit indicator modality.
}
\label{fig:system}
\end{figure*}

\subsection{Role of Social Signals and Error Detection}

Social signals reflect information on robot actions, tasks, and users' mental models of the robot. Treating ``humans as sensors'' allows us to use human behavior to assess robot action~\cite{lewis2009using}. Social signals have already been used in areas such as detecting user confusion~\cite{stiber2024uh} and intention~\cite{belcamino2024gaze}.
When a robot makes an error, users exhibit more social signals, making them reliable error indicators~\cite{cahya2019static, giuliani2015systematic}, 
with error severity influencing responses~\cite{kontogiorgos2021systematic, stiber2020not}.
Common responses include gaze~\cite{severitt2024communication}, 
facial expressions~\cite{stiber2022modeling,bremers2023bystander}, 
verbalizations~\cite{kontogiorgos2021systematic}, and body movements~\cite{giuliani2015systematic}.

Social signals have been modeled to detect and classify conversation failures from social robots~\cite{kontogiorgos2021systematic} and used in virtual reality as an error indicator~\cite{severitt2024communication}.
Human-human interaction data has been modeled for robot error detection using novelty detection but remains scenario-dependent, requiring new data for each context.~\cite{ravishankar2024zero}. In physical interactions, facial AUs have been modeled for real-time error detection and shown to have potential to improve timeliness~\cite{stiber2022modeling,stiber2023using}. However, social signals can be ambiguous without context, leading to reliability issues (\eg false positives)~\cite{stiber2022modeling,severitt2024communication}. 
Prior work has largely relied on post hoc analysis to evaluate performance. To the best of our knowledge, no work has evaluated this approach as part of an interactive robot system. 

\subsection{Proactivity in Error Management}
Prior work has shown that users perceive robots with proactive error mitigation (\eg explanations) as more intelligent compared to those that are reactive~\cite{lemasurier2024reactive}. However, individuals have varied preferences regarding type and level of proactive assistance~\cite{meurisch2020exploring}.
Researchers have explored frameworks for proactive explanations and recovery~\cite{klein2024creating}, considering external factors like task complexity and user privacy, which influence the desired level of automatic recovery~\cite{lee2024rex}. There, however, has been limited research on how users perceive a proactive error detection system when collaborating with a robot. Understanding this is critical for evaluating the effectiveness and acceptance of proactive detection systems in HRI.

\section{Proactive Robot Error Detection System}\label{sec:system}
We developed a proactive, social-signal-based, real-time error detection system for robots. Our goal was to ensure flexibility, timeliness, and reliability, paired with natural voice interactions.
To enable \textit{automatic, proactive} detection, we used social signals---facial action units (AUs) and speech---to identify potential errors before users needed to report them.
To achieve \textit{flexible} detection across tasks and participants without manual adjustments, we leveraged the potential flexibility of social signal models, as demonstrated in previous studies~\cite{stiber2022modeling,stiber2023using}. 
For \textit{reliable} detection, we structured our system inputs using an error-aware conceptual framework~\cite{stiber2023using}, integrating status quo detection inputs (user explicit verbal reporting, robot status, and task information) with social signals. The additional context provided by these inputs improves social signal-based detection, aligning with findings from other domains~\cite{stiber2024uh}. To address potential social signal ambiguity, implicit error indicators are verified through user feedback, reducing false positives~\cite{stiber2022modeling}.

\vspace{-0.35em}

\subsection{System Overview}
Our system (Fig.~\ref{fig:system}) allows verbal interaction and automatically detects errors while performing physical tasks. It handles natural language inputs and uses video of user faces to support two methods of error detection: implicit and explicit. Implicit is system-initiated, using AUs or speech reactions as indicators in a multi-phase process. Explicit is user-initiated via verbal reporting. Once an error is detected, the system starts the process of error mitigation and recovery.

Our system was built on Microsoft's Platform for Situated Intelligence (\textbackslash psi) 
and uses a Kinova Gen3 as the robotic platform, controlled using ROS with a Kinova API on top. It received inputs via a microphone and two Kinect cameras positioned to ensure the user was always in the field of view. 
The system includes components for perception, user intent recognition (verbal commands), two error detection methods (implicit and explicit), error mitigation and recovery, and a robot controller.

\subsection{Perception}
The Perception Component synchronizes input from two cameras and a microphone. The video streams are fed through OpenFace to extract AUs, which are sent to the Implicit Error Detection Component. In parallel, the audio stream is fed through Azure Speech-to-Text and the text is sent to the Intent Recognition Component.  

\subsection{Intent Recognition}
The Intent Recognition Component enables natural interactions by determining, via a large language model (GPT-4), whether the user is requesting the robot to perform an action, explicitly reporting an error, responding to a robot query, verbally reacting to an error, or saying something irrelevant.

\subsection{Implicit Error Detection}
The Implicit Error Detection Component uses social signals and the robot's status (context) to \emph{proactively} detect errors.
This component combines two behavioral signal modalities---speech and AUs---with the robot's actions that were occurring during social signal elicitation (moving, state of the gripper, time since last movement). The detection process has two phases: (1) initial potential error detection using social signals and (2) verification through a robot query. This approach aims to enhance error detection performance while minimizing false positives. Fig.~\ref{fig:teaser} depicts an example interaction workflow and Fig.~\ref{fig:system} features a detailed diagram of this component.

\subsubsection{Phase 1---Initial Potential Error Detection}
Here, a potential error is flagged in one of two ways, depending on the social signal input (described below). These social signals are contextualized with  robot status by checking whether the robot was moving at the time or within the past three seconds, as well as the gripper state. If the potential error occurred while the robot was moving, or within three seconds of when it had been moving, the potential error passes this check. The component then sends a pause command to the Robot Controller, and the system proceeds to phase 2. If it does not pass this check, it is ignored. The phase 1 processes for each input modality (speech and AUs) are outlined below.

\textbf{Speech.}
Implicit detection using speech relies on the Intention Recognition Component to identify potential user reactions to errors. The distinguishing point is that speech-initiated implicit detection occurs when the user is not explicitly reporting an error and the user's speech content lacks information for the LLM to definitively determine that an error occurred (\eg ``you missed it'' or ``drop it''). Instead, the LLM infers that the user might be responding to a possible error and flags it as such.

\textbf{Action Units.}
AUs are fed into an error detection algorithm, trained on the open-source Social Responses to Errors in HRI dataset~\cite{stiber2023using}. The algorithm's structure is similar to the one originally evaluated with that dataset. At each time step, the algorithm first determines if a change in facial expression could be indicative of a potential error and then these classifications are fed into a four-second sliding window. If over half of that window is deemed as a response to a potential error, then a possible error is detected. The threshold for the sliding window can be adjusted based on the user's response in phase 2.

\subsubsection{Phase 2---Verification}
In phase 2, the component verifies with the user that an error has indeed occurred by asking the user a question about how the task is going. This will reduce false positives.
We generate the robot queries using an LLM (GPT-3.5) to ensure the robot asks a variety of yes/no questions (\eg ``Is everything going according to plan?''). 
We ensured that the robot would never explicitly ask if an error had occurred to avoid biasing the user into thinking that the robot is error prone or has false positives.

A user's response to the robot's query can influence the AU modeling algorithm by adjusting the detection threshold. If the user indicates no error has occurred, the threshold is temporarily increased, making the system less sensitive. Over time, the threshold gradually returns to its original level. If the user confirms an error, the threshold remains unchanged.

\subsection{Explicit Error Detection}
The Explicit Error Detection Component detects errors through user verbal reports (\eg ``You made a mistake. You put the nuts in the office box instead of the food box.''), making this component \emph{reactive}.
The error report can be made at any time. Once an error is reported, this component sends a message to the Error Mitigation and Recovery Component, flagging that an error has been detected.

\subsection{Error Mitigation and Recovery}
The focus of our system is the Error Detection Components; therefore, for mitigation and recovery, we had this component trigger an apology and send a recovery command to the robot controller. The recovery behaviors of the robot were pre-programmed.

\subsection{Robot Controller}
The Robot Controller Component handles communication with the Kinova Gen3 robot arm. It relays commands from the intent recognition component to the robot to execute the corresponding action. It receives notifications from the robot when tasks are done, along with robot status information such as gripper position and movement details, and the time since movement, and sends this data to the rest of the system. The Error Detection and Error Mitigation and Recovery Components also communicate with this component to flag the need for the robot to pause, stop, and recover.

\section{Evaluation}\label{sec:eval-study}

\begin{figure}
\includegraphics[width = \columnwidth]{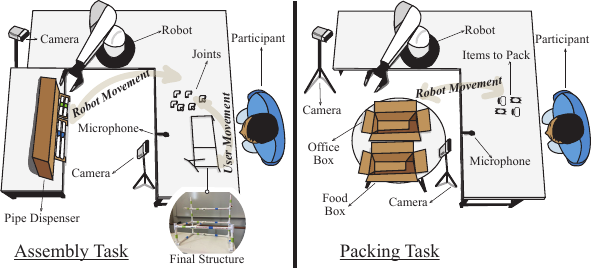}
\caption{Evaluation study setup for assembly and packing tasks.}
\label{fig:study-setup}
\end{figure}

We conducted a user study to evaluate our system and approach. Our evaluation focused on two dimensions: 1) validate that our system can detect errors both reliably and flexibly across tasks and participants and 2) evaluate our system error detection performance (timeliness) and user preference and perception of the robot as a teammate compared with a \emph{reactive} detection system. 
In a between-subjects design, participants were randomly assigned to either our \emph{proactive} error detection robotic system or a \emph{reactive} one. We varied only the detection components, keeping the rest of the robotic system unchanged for a direct detection comparison.

\begin{itemize} [leftmargin=*]
    \item \emph{Reactive Error Detection.} In this condition, the system relied solely on users manually and explicitly 
    reporting errors, employing only the explicit error detection component. It detected errors using a modified intention recognition LLM prompt,
    allowing participants to report errors in any wording that was natural to them. We chose this as our comparison because manual reporting is commonly used in deployed robots (\eg~\cite{edirisinghe2024field}). Additionally, this enables a controlled comparison between a system with implicit social signals (proactive) for error detection and one without (reactive) with results reflecting the direct impact of incorporating implicit social signals. The reactive condition is also human baseline illustrating users' natural performance for detecting errors. Prior work has shown that social signals have the potential for faster detection than human reporting~\cite{stiber2023using}.
    
    \item \emph{Proactive Error Detection.} This condition employed our proactive system (described in Section~\ref{sec:system}, using both the explicit and implicit detection components).
\end{itemize}

\subsection{Study Tasks and Error Manipulations}

We evaluated our system over two tasks: assembly and packing (Fig.~\ref{fig:study-setup}).
The AU-based error detection ML model was trained on AUs exhibited during an assembly task (albeit with a different structure and setup), but not on any data from a packing task. This allowed us to assess whether our proactive system was flexible enough to detect errors in a previously unseen task. 
To use the systems for both tasks, we modified the intent recognition component for both systems, prompting them specifically for each task.
Both tasks had two pre-programmed errors that were unexpected for the participant. However, since both systems were autonomous, participants could potentially experience other errors.

\textbf{Assembly Task.} Participants built a bench with the robot, using provided joints and a photo of the structure. Participants could request pipes from the robot verbally in any format, as long as they specified the pipe color. The assembly task included a physical error (failing to grasp a pipe) and a conceptual error (grasping an incorrect pipe). To ensure participants had time to develop a mental model of the robot and mitigate the novelty effect, errors were set to happen only after a minimum of three pipe requests, with timing dependent on how the participant constructed the structure.

\textbf{Packing Task.} Contextualized as a warehouse packing task, this task had participants verbally instruct the robot on which objects should be placed in which boxes. There were two categories of objects corresponding to two boxes: office supplies and food. Participants could instruct the robot in any way and order they chose. The packing task included 
a conceptual error (placing a nuts jar into a box designated for office supplies) and a physical error (pausing for seven seconds between the two boxes while holding earbuds). The physical error might not be perceived as an error by every participant and was included to see if the system could detect the error based on participants' perceptions.

\subsection{Measures}

\subsubsection{Manipulation Check}
Participants were asked if the robot made errors (binary choice). Those reporting no errors were excluded. We also asked participants to report the number of errors the robot made and to describe the errors, allowing us to include errors that were not pre-planned.

\subsubsection{Objective Measures}
The values calculated below were derived from the logs generated by the systems.

\textbf{Error Detection Delay} (seconds). This evaluates timeliness as the time between error occurrence (denoted by the robot's trajectory and gripping) and error detection. For implicit detection, it depends on the triggering social signal: speech detection starts when intent recognition identifies a verbal reaction, and AU detection begins when the AU model signals a potential error. For explicit detection, time starts when the participant reports the error.
    
\textbf{Percent of Errors Detected}. The percentage of user self-reported observed errors successfully detected by the system.

\subsubsection{Subjective Measures}
We relied on questionnaires to understand participants' evaluation of the robot's error handling and their perceptions of the robot as a teammate.

\textbf{Error Handling Satisfaction} (scale: 1--7). This is a two-item scale (Cronbach's $\alpha = 0.81$) created to assess participants' satisfaction with how the robot managed errors.

\textbf{Robot Teammate Quality} (scale: 1--7). This is a nine-item scale (Cronbach's $\alpha = 0.93$) used to evaluate participants' perception of how good of a teammate the robot was.

\begin{figure*}
\includegraphics[width = \textwidth]{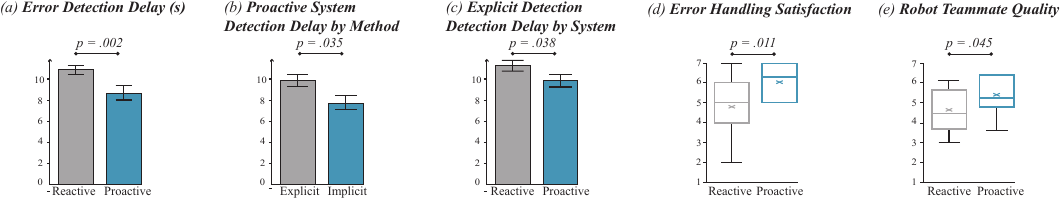}
\caption{\textbf{Assembly} Results: bars depict mean; error bars are standard error. (a) Impact of system type on detection timeliness (error detection delay). (b) Detection methods' impact on error detection delay for the \emph{proactive} system. (c) Impact of  system type on error detection delay for explicit detection. Participants' perception of systems' error handling and robot. Cross is mean; box shows quartiles; line shows median. (d) Impact of system type on satisfaction with error handling. (e) Impact of system type on perception of robot as a teammate.}
\label{fig:results-assembly}
\end{figure*}

\begin{figure}
\includegraphics[width = \columnwidth]{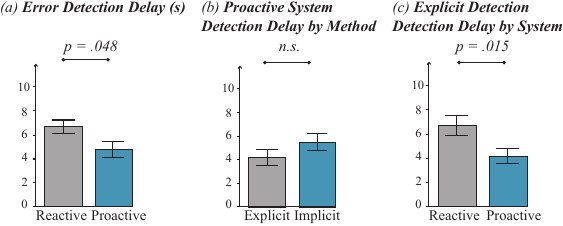}
\caption{\textbf{Packing} Results: bars depict mean; error bars are standard error. (a) Impact of system type on detection timeliness (error detection delay). (b) Detection methods' impact on error detection delay for the \emph{proactive} system. (c) Impact of  system type on error detection delay for explicit detection.}
\label{fig:results-packing}
\end{figure}

\subsection{Study Procedure} \label{sec:procedure}
After consenting to the study, participants were instructed on how to interact with the robot and completed a practice task to familiarize themselves with the system.
After the practice task and once participants verbally confirmed their understanding, the experimenter initiated the assembly task and exited the room. The participants executed the task 
and 
were then asked to fill out questionnaires regarding their experience and perceptions of interacting with the robot. They were then informed they would interact with a different robot system in a new context, to reset their mental model.
Participants then repeated this process for the packing task. After both rounds, participants
provided their demographic information and the experimenter conducted a semi-structured interview. The study was approved by our institutional review board and lasted, on average, 50 minutes. Participants were compensated US\$15.

\subsection{Participants}\label{sec:participant}
Thirty participants were recruited for this evaluation; two participants were excluded due to manipulation check failure and the speech-to-text portion of the system not transcribing reasonably. Additionally, four trials were excluded due to equipment failure (\ie LLM service went down, pipe dispenser broke). Therefore, the results reported below are based on 28 participants (18 female, 10 male), aged 19 to 55 ($M = 26.5, \mathit{SD} = 8.04$). Of those participants, 14 were in the proactive detection and 14 in the reactive detection condition. 
Participants had moderate prior technology experience ($M = 3.78, \mathit{SD} = 1.02$, 5-point scale) and little prior robot experience ($M = 2.22, \mathit{SD} = 1.14$, 5-point scale).

\section{Results}

We used the Shapiro-Wilk test to assess our measures' normality. Error detection delay, which was normally distributed, was analyzed using a two-tailed t-test. 
The subjective metrics, which were not normally distributed, were analyzed using the Mann-Whitney U test. 
We note that 
\emph{tasks} were intended merely to evaluate the systems in different contexts for detection performance, not as a manipulation. Results are reported separately without statistical comparisons between tasks. Additionally, the subjective metrics were only reported for the assembly task as the purpose of the packing task was just to evaluate our system's detection flexibility. See Fig.~\ref{fig:results-assembly} for detection performance and user perception results for the assembly tasks and Fig.~\ref{fig:results-packing} for detection performance results for the packing task.

\vspace{-0.3em}

\subsection{Assembly Task}

\subsubsection{Error Detection Delay}
We observed a significant effect of system condition on error detection delay, $t(54) = 3.25, p = .002, d = 0.88$. The \emph{proactive} system ($M = 8.98, \mathit{SD} = 2.55$) detected errors with less delay compared to the \emph{reactive} system ($M = 11.22, \mathit{SD} = 2.54$), as shown in Fig.~\ref{fig:results-assembly}a. 
Focusing on the detections within the \emph{proactive} system condition, we found a statistically significant difference in error detection delay between the Explicit and the Implicit Error Detection Components, $t(26) = 2.25, p = .035, d = 0.87$. Implicit detections ($M = 7.70, SD = 2.14$) were faster than  explicit ones ($M = 9.77, SD = 2.51$), as illustrated in Fig.~\ref{fig:results-assembly}b.
Additionally, when analyzing just the Explicit Detection Component  between systems, we found a statistically significant difference in error detection delay, $t(42) = 1.84, p = .038, d = 0.57$ (Fig.~\ref{fig:results-assembly}c). Detections using the explicit component in the \emph{proactive} system ($M = 9.77, \mathit{SD} = 2.51$) were faster than in the \emph{reactive} system ($M = 11.22, \mathit{SD} = 2.54$).

\subsubsection{Percent of Errors Detected}
All errors for both systems were detected. In the \emph{proactive} system, 38.5\% of errors were detected using the implicit detection component, comprising of AUs (60\%) and speech (40\%).

\subsubsection{Error Handling Satisfaction}
Looking at the perception of how well  errors were handled, we found a significant effect of the detection system, $U = 35, p = .011, r = 0.50$ (Fig.~\ref{fig:results-assembly}d). The \emph{proactive} detection system ($\mathit{Mdn} = 6.25$) was perceived as better than the \emph{reactive} one ($Mdn = 5$).

\subsubsection{Robot Teammate Quality}
We found a medium effect of the detection system on participants' perception of how good the robot was as a teammate ($U = 44.50, p = .045, r = 0.39$); specifically, the robot with the \emph{proactive} detection system ($\mathit{Mdn} = 5.22$) was viewed as a better teammate than the one with the \emph{reactive} system ($\mathit{Mdn} = 4.44$), see Fig.~\ref{fig:results-assembly}e.

\subsection{Packing Task}
\subsubsection{Error Detection Delay}
The \emph{proactive} detection system ($M = 4.78, SD = 1.82$) had statistically significant less delay than the \emph{reactive} one ($M = 6.74, SD = 2.79$), $t(46) = 2.11, p = .048, d = 0.83$ (Fig.~\ref{fig:results-packing}a). However, there was no statistically significant difference between explicit ($M = 4.04, SD = 1.54$) and implicit detection ($M = 5.44, SD = 1.89$) for detections done in the \emph{proactive} system condition, $t(24) = 1.47, p = .17, d = 0.81$ (Fig.~\ref{fig:results-packing}b). Looking at detection delay difference between explicit detections done with the \emph{proactive} and \emph{reactive} systems, the difference was statistically significant, $t(32) = 2.71, p = .015, d = 0.81$. Explicit detection in the \emph{proactive} condition ($M = 4.04, \mathit{SD} = 1.54$) was faster than the \emph{reactive}'s ($M = 6.74, \mathit{SD} = 2.79$), see Fig.~\ref{fig:results-packing}c.

\subsubsection{Percent of Errors Detected}
The \emph{reactive} system detected 91.7\% of the errors correctly, whereas the \emph{proactive} system resolved 100\% of the errors. In addition, the \emph{proactive} system detected 53.8\% of the errors it encountered using the implicit social signal error detection component through AUs (85.7\%) and speech (14.3\%).

\section{Discussion}
We developed a proactive error detection system using implicit and explicit social signals as indicators of robot errors.
Our evaluation revealed that our system can detect robot errors \emph{automatically}, \emph{reliably}, and \emph{flexibly}, and is more \emph{timely} and viewed more favorably than a reactive method---highlighting its advantages over the status quo.

\subsection{Our Proactive Robotic System is Reliable and Flexible}
This is demonstrated across both assembly and packing tasks with multiple participants. 
Our system successfully detected 100\% of errors in both tasks, illustrating \textbf{reliability}. As one participant (P7, proactive) noted: \pquotes{``I wasn't planning on reporting it because I wasn't sure it would be able to fix it. But then, it asked if everything was going well, and I said no. And then it was able to correct its own mistake...''.} Detections used both implicit and explicit components, with most implicit detections relying on AUs.
Notably, in the packing task, most detections were made using the implicit component, primarily AUs, despite the AU model not being trained on similar tasks, highlighting the technique's adaptability across tasks. It also detected errors using the implicit component for first-time users, highlighting the potential \textbf{flexibility} of our technique for other tasks and users.

\vspace{-0.3em}

\subsection{Benefits of Proactive Error Detection via Social Signals}\label{sec:benefits-discussion}

\subsubsection{Better Error Detection Performance}
The \emph{proactive} system detects errors faster than the \emph{reactive} system.
In the assembly task, the \emph{proactive} system detected errors $2.24s$ faster than the \emph{reactive} system, with the implicit component identifying errors $2.07s$ faster than explicit. 
This aligns with expectations from prior work~\cite{stiber2023using} that detection using social signals can capitalize on reflexive behaviors exhibited earlier than intentional reporting. Our detection system proved to be more \textbf{timely} than the reactive one.

\subsubsection{Better User Perception and Preference}
Participants were more satisfied with the robot's error handling in the \emph{proactive} condition than the \emph{reactive} one. P24 noted, \pquotes{``He [the \textbf{reactive} robot] \textbf{wasn't} like \textbf{aware} that he made a mistake. So I had to tell him,''} reinforcing the perception that the \emph{reactive} system lacked error awareness.
In the assembly task, participants rated the robot with the \emph{proactive} system more favorably as a teammate than the \emph{reactive} one.
In fact, P1 (proactive) mentioned, \pquotes{``...it [the robot] was asking in between whether the task was going well, which made me feel like it's a teammate.''}

The majority of \emph{proactive} condition participants (64.3\%) preferred an error detection system similar to what they experienced, combining user-initiated and robot-initiated error detection. P10 (proactive) unprompted stated that their ideal detection system would be \pquotes{``...a mixture of the robot reading physical cues from the operator and the robot having internal checks each time they performed a task...''} Conversely, the majority of \emph{reactive} condition participants (84.6\%) did not want the detection system they experienced.

\vspace{-0.3em}

\subsection{Proactive Interaction Shaped User Behavior}\label{sec:proactiveShape}
Beyond perception and preference, proactive detection influenced participants' behavior and shaped how they reported errors. 
The robot \pquotes{``seemed like it was only asking when it was unsure about what it had done,''} which could have made the robot appear to have greater awareness to participants---turning a potential interaction breakdown (false positives) into a positive behavior. 
For both tasks, participant error reportings 
were more prompt during interactions with the \emph{proactive} system than the \emph{reactive} one. 
Proactivity appeared to influence participants to report errors faster.

\subsection{Recommendations for Achieving Robot Error Awareness}\label{sec:recommendations}
We propose six recommendations to inform future development of robot error awareness. 

\highlight{lightblue}{\textbf{Recommendation \#1} (Collaborative Error Detection)} \emph{Error aware systems should allow multiple detection techniques: robot-initiated and user-initiated detection.}

\noindent Just as common user interfaces offer multiple ways to perform the same task for increased accessibility,
error detection systems should support multiple methods for identifying errors, as users' reactions and tendencies vary.
The majority of participants, regardless of the system they interacted with, wanted an error detection system to allow both human and robot active involvement ($n=16$).
Since participants valued both user and system involvement, combining these in a collaborative manner could enhance detection speed, reliability, and interaction perception.

\highlight{lightblue}{\textbf{Recommendation \#2} (Human-in-the-loop)} \emph{Error aware systems should involve users even in the robot-initiated detection process (\eg for verification).}

\noindent Involving users in robot-initiated error detection via robot queries (error verification) improves reliability, maintains proactivity, and is viewed favorably by participants. Given that behavioral signals can be noisy and ambiguous~\cite{stiber2022modeling},
these queries helped prevent false positives.

\highlight{lightblue}{\textbf{Recommendation \#3} (Mitigation through Proactivity)} \emph{Error aware systems should use proactivity and proactive behavior, such as robot-initiated queries, to ``preemptively mitigate'' potential negative effects of errors.}

\noindent Interestingly, the queries triggered by initial detection false positives---typically seen as detrimental to an interactive system---were perceived positively by participants. They found this approach helpful for managing errors and maintaining engagement with the robot. The queries offered an opportunity for the system to help users better understand the robot's capabilities and limitations, potentially turning an interaction breakdown into a constructive experience. 

\highlight{lightblue}{\textbf{Recommendation \#4} (Learning from Human Feedback)} \emph{Error aware systems should use user error reports, feedback, and query answers to help them learn from the erroneous representation and misclassifications of human behavior.} 

\noindent When reporting errors and providing feedback, participants often volunteered additional information beyond what was needed for the report or to answer the question (\eg P11 stated \pquotes{``That should have went in the food box not the office box.''}), which could be useful to assist in recovery or enable the robot to learn from its incorrect detections, improving future performance.
This could support techniques to reduce ambiguity and refine behavioral models through interactivity.

\highlight{lightblue}{\textbf{Recommendation \#5} (Flexible Reporting and Feedback)} \emph{Error aware systems should support flexible interactions, offering users various direct and indirect ways to report and provide feedback, accommodating communication styles.}

\noindent A few participants in the study expressed discomfort in directly reporting errors made by the system or responding to  the robot's binary question. For example,
P9 said \pquotes{``I feel uncomfortable saying that it did something wrong... So when it detected that it made a mistake if I felt a lot of relief...''} Not all users may feel comfortable engaging in direct communication. Our system addressed this by integrating an LLM to allow  participants to report in any manner they chose. 
However, carefully framing feedback questions is important, as it can affect user perceptions and behavior~\cite{candon2023verbally}.

\highlight{lightblue}{\textbf{Recommendation \#6} (Adaptive Proactivity)} \emph{Error aware systems should adjust proactivity level based on the task, error context, nature of the error, and user preferences.}

\noindent Participants want different levels of proactivity in error management, depending on the nature of the tasks and errors, error handling performance, and ability to provide collaborative assistance. Interviews revealed that they wanted the robot to have a lower level of proactivity when the task had high stakes, task or error recovery was complex, or the task was service related or naturally involved a user ($n = 9$). In those cases, the suggestion was for the robot to include the user by asking for help or providing explanations, and then walking through the correction plan ($n = 5$).
f the task is too dangerous for user involvement, users prefer full robot autonomy. This is consistent with findings from a prior study on robot error explanations and repair~\cite{lee2024rex}.

\subsection{Limitations and Future Work}
Our system was evaluated over a single interaction. Longer-term interactions or multiple sessions could impact the effectiveness of implicit error detection, as behavioral signals may change as users' mental models of the robot evolve. Additionally, we do not draw conclusions about task differences, but the timeliness of detections did vary and may be due to task context (in one task, the user collaborates and is not always focused on the robot, while in the other, the user is idle). Future work should explore how context in which an error occurs affects user reactions and error detections.
Another extension is to use the robot's query responses for continual learning to improve implicit error detection models. Currently, the system only uses query responses to adjust AU model's thresholds. Future work should explore using user error reports and query answers for model fine-tuning, adaptation, or continual learning in longer-term interactions.




\balance

\bibliographystyle{IEEEtran}
\bibliography{references}

\end{document}